\documentclass{article}
\usepackage{booktabs}
\usepackage{spconf,graphicx}
\usepackage{bm} 
\usepackage{amsmath}
\usepackage{bbding}
\usepackage{array}
\usepackage{amssymb}
\usepackage{multirow}

\title{LOGO-Former: Local-Global Spatio-Temporal Transformer for Dynamic Facial Expression Recognition}
\name{Fuyan Ma, Bin Sun, Shutao Li\thanks{This work is supported by the National Natural Science Fund of China (62171183, 62221002), the Hunan Provincial Natural Science Foundation of China (2022JJ20017), and partially sponsored by CAAI-Huawei MindSpore Open Fund.	
Corresponding author: Bin Sun.}}
\address{College of Electrical and Information Engineering, Hunan University, Changsha, China}

\begin{document}
%
\maketitle
\begin{abstract}
Previous methods for dynamic facial expression recognition (DFER) in the wild are mainly based on 
Convolutional Neural Networks (CNNs), whose local operations ignore the long-range dependencies in videos.
Transformer-based methods for DFER can achieve better performances but result in higher FLOPs and computational costs.
To solve these problems, the local-global spatio-temporal Transformer (LOGO-Former) is proposed to capture discriminative features within each frame and model contextual relationships among frames while balancing the complexity.
Based on the priors that facial muscles move locally and facial expressions gradually change, we first restrict both the space attention and the time attention to a local window to capture local interactions among feature tokens. Furthermore, we perform the global attention by querying a token with features from each local window iteratively to obtain long-range information of the whole video sequence.
In addition, we propose the compact loss regularization term to further encourage the learned features have the minimum intra-class distance and the maximum inter-class distance.
Experiments on two in-the-wild dynamic facial expression datasets (i.e., DFEW and FERV39K) indicate that our method provides an effective way to make use of the spatial and temporal dependencies for DFER.
\end{abstract}
\section{Introduction}
\label{sec:intro}

Facial expression recognition (FER) has been an emerging topic, due to its essential real-world applications in driver safety monitoring, human robot emotional interaction, elderly healthcare and so on.
Previous intensive studies (such as \cite{zhao2021learning,ma2021facial}) have been conducted on static facial expression recognition (SFER). However, a facial expression is a dynamic process, which consists of various facial muscle motions in different facial regions.

Previous dynamic facial expression recognition (DFER) methods can be mainly divided into two categories (i.e., static frame-based methods and dynamic sequence based methods) \cite{li2020deep}.
Most of the static frame-based methods utilize Gabor wavelets \cite{lee2016collaborative}  and convolutional features\cite{yang2018facial,liu2018conditional} to select peak (apex) frames in videos, and further conduct facial expression recognition on these frames.
Although these methods perform well by selecting peak frames, they neglect the temporal dynamics and correlation among facial frames.
Different from static frame-based methods, dynamic sequence based methods usually use 3D convolution neural networks (3DCNN) \cite{ayral2021temporal}, long-short term memory (LSTM) \cite{vielzeuf2017temporal} to learn the spatio-temporal relationships, which can model long-term dependencies and improve the performance of DFER.
The performances of these methods are still far from being satisfactory, because of occlusions, variant head poses, poor illumination and other unexpected issues in real-world scenes.

Recent flourishing of Transformers on computer vision tasks has considerably deepened our understanding about discriminative feature representation and contextual information modeling.
For example, Li \textit{et al.} \cite{li2021self} exploit the bidirectional Transformers to capture the temporal information among frames.
Directly extending the vanilla Transformer \cite{dosovitskiy2020image, ma2022spatio} for DFER needs to perform the multi-head self-attention jointly across all $S$ spatial locations and $T$ temporal locations.
Namely, the full space-time attention that has complexity $O(T^2S^2)$ puts heavy computational burdens within the vanilla Transformer framework for efficient dynamic facial expression recognition.
A simple solution to reduce the cost of full space-time attention is to calculate spatial-only attention followed by temporal averaging, which has complexity of $O(TS^2)$.
Very recently, the divided space-time attention \cite{bertasius2021space} and the space-time mixing attention \cite{bulat2021space} have been proposed to induce significant computational overheads compared to the full space-time attention, which reduce the complexity to $O(TS^2 + T^2S)$ and $O(TS^2)$, respectively.

\begin{figure}[t]
	\centering
	\includegraphics[width=\linewidth]{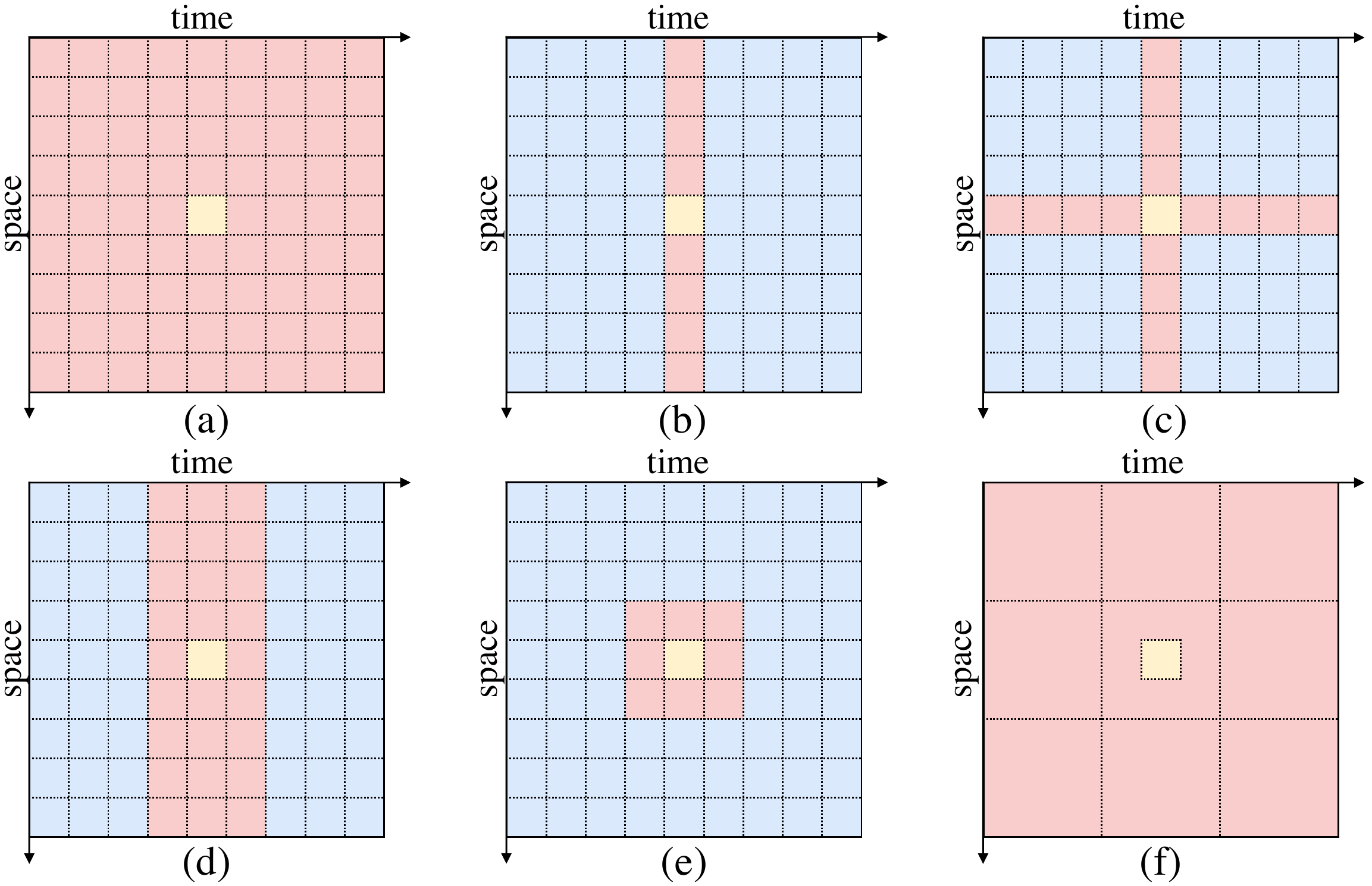}
	\caption{Visualization of different space-time self-attention schemes. For better illustration, we denote the query token in light yellow, the key tokens in rose and non-attentive tokens in light blue. (a) Full space-time attention \cite{dosovitskiy2020image}: $O(T^2S^2)$. (b) Spatial-only attention: $O(TS^2)$. (c) Divided space-time attention \cite{bertasius2021space}: $O(TS^2 + T^2S)$. (d) Space-time mixing attention \cite{bulat2021space}: $O(TS^2)$. (e) and (f): Our local and global space-time attention. 
	}
	\label{intro}
  \end{figure}

Different facial muscles move within local facial regions, and facial expressions gradually change within adjacent frames in a video.
Our aim is to exploit the spatio-temporal information present in videos while minimizing the computational costs of Transformers for efficient dynamic facial expression recognition.
To achieve this, the LOcal-GlObal spatio-temporal Transformer (LOGO-Former) is proposed to capture short- and long-range dependencies, and meanwhile reduce the computational costs of Transformers.
We compute the self-attention within non-overlapping windows to capture local interactions among tokens.
Such local space-time attention fails to capture global information.
Therefore, the global space-time attention is utilized to make the query token attend to window-level tokens, as shown in Fig.\ref{intro}.
To further increase the model discriminant ability, we propose the compact loss regularization term to decrease the intra-class distance and increase the inter-class distance.
The quantitative results and the visualization results demonstrate the effectiveness of our method for in-the-wild dynamic facial expression recognition.


\begin{figure*}[t]
	\centering
	\includegraphics[width=0.9\textwidth]{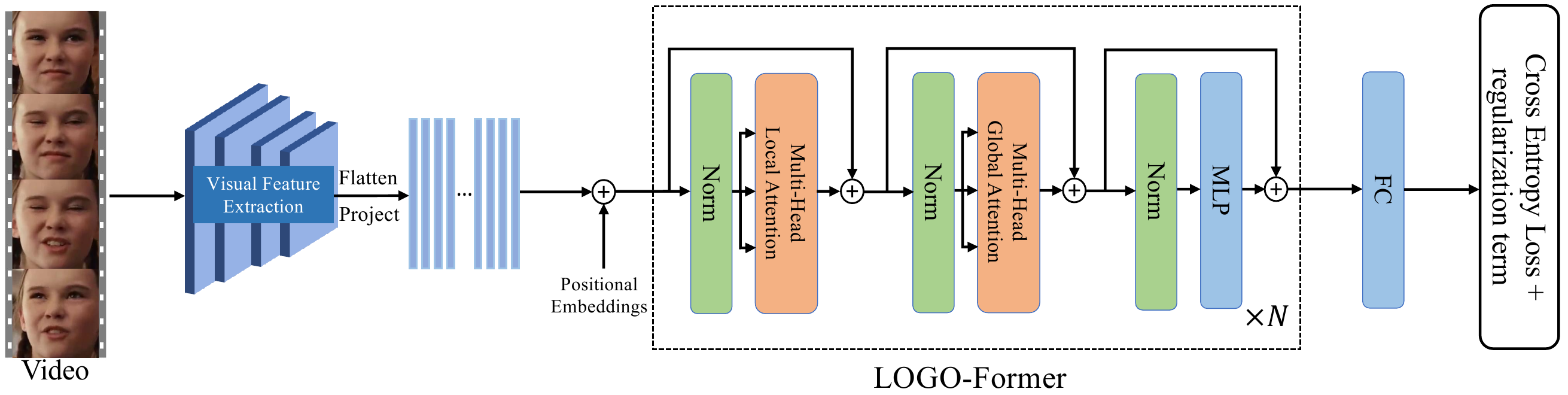}
	\caption{Overview of our method for dynamic facial expression recognition. The facial frames sampled from a video are processed by the CNN backbone to obtain frame-level feature sequences. The LOGO-Former jointly calculates the spatial attention and the temporal attention for capturing discriminative feature tokens.
	}
	\label{STT}
  \end{figure*}

\section{THE PROPOSED METHOD}
\label{method}

\subsection{Input Embedding Generation}
The proposed framework is shown in Fig. \ref{STT}, which consists of three parts, i.e., input embedding generation, LOGO-Former and the compact loss supervision.
Given an image sequence $\bm{X} \in \mathbb{R}^{F\times H_{0}\times W_{0}\times 3}$ with $F$ RGB facial frames of size $H_{0}\times W_{0}$ sampled from the video, we utilize the CNN backbone to extract frame-level features.
A standard CNN backbone (e.g., ResNet18) is used to generate high-level feature maps of size $H \times W$ for each frame. The clip-level features $\bm{f}_{0} \in \mathbb{R}^{F\times H \times W \times C}$ are obtained by concatenating all frame-level feature maps.

Subsequently, we flatten the spatial dimension of the clip-level feature maps $\bm{f}_{0}$ and project them by a $1 \times 1$ convolution, resulting in a new feature sequence $\bm{f}_{1} \in \mathbb{R}^{F\times (H\times W)\times d}$.
It is noted that the temporal order of $\bm{f}_{1}$ is in accordance with that of the input $X$.
To supplement the spatio-temporal positional information for the feature sequence, we incorporate the learnable positional embeddings with $\bm{f}_{1}$.
We also prepend the classification token $[CLS]$ to the sequence at the temporal dimension, which models the global state of the sequence and is further used for recognition.
Similarly, the temporal positional embedding is also added. Finally, the input embedding $\bm{X}^{0}$ to the spatio-temporal Transformer is obtained.

\subsection{LOGO-Former}

The LOGO-Former consists of $N$ blocks and each block is composed with the multi-head local attention and the multi-head global attention, which iteratively learns the contextual and discriminative spatio-temporal feature representation.
\par
{\bfseries Multi-Head Local Attention:}
Given that nearby tokens have stronger corrections than no-local tokens, we perform multi-head local attention within non-overlapping window to model local interactions among these tokens.

Taking the input feature map with $F\times H\times W$ (we omit the $[CLS]$ token here for simplicity) as the input, we evenly split it into several windows with size of $f \times hw$, leading to $\frac{F}{f} \times \frac{HW}{hw}$ windows, as shown in Fig.\ref{intro}.
We flatten these tokens within a window $(i,j)$, which can be denoted as $\bm{X}_{i,j} \in \mathbb{R}^{(fhw)\times d}$.
The multi-head local attention of the $k$-th block is formulated as
\begin{equation}
	\label{eq3}
	\bm{Y}^{k} = \bm{X}^{k-1}_{i,j} + {\rm{MSA}}({\rm{LN}}(\bm{X}^{k-1}_{i,j})),
\end{equation}
where ${\rm{LN}}$, ${\rm{MSA}}$ represent layer normalization and multi-head self-attention, respectively.
The residual shortcut is used after every attention block to avoid gradient vanishing.
As we restrict the attention calculation within local windows, the complexity of our multi-head local attention is $O(FHWfhw)$, where the cost is computed as
\begin{equation}
	\label{eq5}
	O({\rm{MHLA}}) = (fhw)^2\times \frac{FHW}{fhw} = FHWfhw.
\end{equation}
Compared with the full space-time attention with $O(T^2S^2)$, which can also be rewritten as $O(F^2H^2W^2)$, our local attention is much more efficient.

{\bfseries Multi-Head Gloabl Attention:}
Although our local attention is computationally efficient, it lacks the ability of capturing global information.
To further capture the global correlation across the frame sequence, we apply the multi-head global attention as a complement for learning long-range dependencies.

Similar with the vanilla multi-head self-attention mechanism, our global attention also takes a query $\bm{Q}$, a key $\bm{K}$ and a value $\bm{V}$ as the inputs.
To lessen the computational overheads, we propose to down-sample the inputs $\bm{K}$ and $\bm{V}$ by the window-wise pooling.
Specifically, we apply the convolution operation to separating and pooling the feature maps into non-overlapping regions, where each region is a spatio-temporal abstract of the feature map. 
Each region is used to pass global contextual information to each query token.
The multi-head global attention is formulated as
\begin{equation}
	\bm{Q}^{k} = W^{k}_{Q}{\rm{LN}}(\bm{Y}^{k}),
\end{equation}
\begin{equation}
	\bm{K}^{k} = W^{k}_{K}{\rm{LN}}({\rm{Pool}}(\bm{Y}^{k})),
\end{equation}
\begin{equation}
	\bm{V}^{k} = W^{k}_{V}{\rm{LN}}({\rm{Pool}}(\bm{Y}^{k})),
\end{equation}
\begin{equation}
	{\rm{MHGA}}(\bm{Y}^{k}) = \bm{Y}^{k} + {\rm{softmax}}(\frac{\bm{Q}^{k}(\bm{K}^{k})^T}{\sqrt{D_{h}}})\bm{V}^{k},
\end{equation}
\begin{equation}
	\bm{X}^k = {\rm{MHGA}}(\bm{Y}^{k}) + {\rm{MLP}}({\rm{LN}}({\rm{MHGA}}(\bm{Y}^{k}))),
\end{equation}
where ${\rm{Pool}}$ denotes the window-wise pooling method (e.g. average pooling, convolution), ${\rm{softmax}}$ represents the softmax function, and $D_{h}$ is the hidden dimensions for each head.
In addition, MLP denotes a multi-layer perceptron for non-linear transformation and the residual shortcut is also used here.
Without loss of generality, suppose that we pool the feature maps $\bm{Y}_k$ into $\frac{FWH}{fwh}$ tokens, the complexity of our multi-head global attention is
\begin{equation}
	O({\rm{MHGA}}) = (\frac{FWH}{fwh}) FWH = \frac{(FHW)^2}{fwh}.
\end{equation}

Therefore, the whole attention complexity of our LOGO-Former can be computed as
\begin{equation}
	\begin{aligned}
		O({\rm{Ours}}) &= FHWfhw + \frac{(FHW)^2}{fwh} \\
		& < \underbrace{(HW)^{2}F+HWF^2}_{{\rm{TimeSformer}}}  < \underbrace{(FHW)^2}_{{\rm{MSA}}}.\\
	\end{aligned}
\end{equation}
Finally, we apply a single fully connected (FC) layer to the classification token $\bm{X}^{N}_{(0,0)}$ of the final block:
\begin{equation}
	\bm{p} = FC(\bm{X}^{N}_{(0,0)}),
\end{equation}
where $\bm{p}$ is the prediction distribution of $C$ facial expression classes.

\begin{table*}[t]
	\caption{Comparison with other state-of-the-art methods on DFEW with 5-fold cross validation. The best results are in bold. \underline{Underline} represents the second best. TI denotes time interpolation and DS denotes dynamic sampling. The evaluation metrics include the unweighted average recall (UAR) and the weighted average recall (WAR).}
  \centering
	\resizebox{0.95\textwidth}{!}{

  \begin{tabular}{c|c|ccccccc|cc}
		\toprule

	\multirow{2}{*}{Method} & Sample & \multicolumn{7}{c|}{Accuracy of Each Emotion (\%)}  & \multicolumn{2}{c}{Metrics (\%)} \\
	\cmidrule{3-11}
	&  Strategies     & Happiness & Sadness & Neutral & Anger & Surprise & Disgust & Fear  & UAR  & WAR    \\
	\midrule
	R(2+1)D18                & TI                                                                           & 79.67     & 39.07   & 57.66   & 50.39 & 48.26    & \underline{3.45}    & 21.06 & 42.79           & 53.22           \\
	ResNet18+LSTM         & TI                                                                           & 78.00     & 40.65   & 53.77   & 56.83 & 45.00    & \textbf{4.14}    & 21.62 & 42.86           & 53.08            \\
	3D R.18+Center Loss     & TI                                                                           & 78.49     & 44.30   & 54.89   & 58.40 & 52.35    & 0.69    & 25.28 & 44.91           & 55.48             \\
	EC-STFL                  & TI                                                                           & 79.18     & 49.05   & 57.85   & 60.98 & 46.15    & 2.76    & 21.51 & 45.35           & 56.51            \\
	3D Resnet18           & DS                                                                           & 76.32     & 50.21   & 64.18   & 62.85 & 47.52    & 0.00    & 24.56 & 46.52           & 58.27           \\
	ResNet18+LSTM           & DS                                                                           & 83.56     & 61.56   & \underline{68.27}   & 65.29 & 51.26    & 0.00    & 29.34 & 51.32           & 63.85             \\
	Resnet18+GRU           & DS                                                                           & 82.87     & \underline{63.83}   & 65.06   & 68.51 & 52.00    & 0.86    & 30.14 & 51.68           & 64.02            \\
	Former-DFER          & DS                                                                           & \underline{84.05}     & 62.57   & 67.52   & \underline{70.03} & \textbf{56.43}    & \underline{3.45}    & \underline{31.78} & \underline{53.69}           & \underline{65.70}               \\
  \midrule
	\textbf{LOGO-Former}              & DS    &  \textbf{85.39}         &   \textbf{66.52}      &  \textbf{68.94}       &  \textbf{71.33}     &  \underline{54.59}        & 0.00        &  \textbf{32.71}     &    \textbf{54.21}     &  \textbf{66.98}              \\
\bottomrule

\end{tabular}
  }
\label{dfew}
\end{table*}

{\bfseries Compact Loss Regularization:} Learning discriminative spatio-temporal features for in-the-wild DFER requires the loss function to have the ability of maximizing the feature distance between different categories. To achieve this, we propose to use the symmetric Kullback-Leibler (KL) divergence $\{\mathcal{D}(\bm{u'}||\bm{p'})\ + \mathcal{D}(\bm{p'} ||\bm{u'})\}$ to measure the difference between the distributions $\bm{u'}$ and $\bm{p'}$ and impose the constraint on the prediction distribution $\bm{p'}$, where $\bm{u'}$ is the uniform distribution over $C-1$ and $\bm{p'}$ is the prediction distribution, but excluding the probability of the corresponding target $y$. $\bm{u'}$ is calculated by the softmax function:
\begin{equation}
	\bm{p'} = {\rm{softmax}}(\{\hat{y}_c^{logits}\}_{c\neq y}),
\end{equation}
where $\{\hat{y}_c^{logits}\}_{c\neq y}$ represents the non-target predicted logits.
Therefore, the regularization term is formulated as
\begin{small}
\begin{equation}
	\begin{aligned}
		&\mathcal{L}_{term} = \mathcal{D}(\bm{u'}||\bm{p'})\ + \mathcal{D}(\bm{p'} ||\bm{u'})\\
		&=\sum_{c \neq y}\frac{1}{C-1}\log\big(\frac{1}{(C-1)\bm{p'}}\big) + \sum_{c \neq y}\log\big(\frac{1}{(C-1)\bm{p'}}\big).\\
	\end{aligned}
\end{equation}
\end{small}
With the regularization term $\mathcal{L}_{term}$, the cross entropy loss guides the model to increase the uniformity of the non-target predicted logits.

\section{Experiments}
\label{experiments}


We use two in-the-wild DFER datasets (i.e., DFEW \cite{jiang2020dfew} and FERV39K \cite{wang2022ferv39k}) to evaluate our proposed method.
For both DFEW and FERV39K, the processed face region images are officially detected, aligned and publicly available.
Our model is trained with the batch-size of 32 on DFEW and FERV39K for 100 epochs with two NVIDIA GTX 1080Ti GPU cards.
The SGD optimizer \cite{ruder2016overview} with an initial learning rate of 0.001 and sharpness-aware minimization \cite{foret2020sharpness} are used to optimize our proposed model. 
We use the pretrained ResNet18 on MS-Celeb-1M \cite{guo2016ms} as our CNN backbone.
The number of spatio-temporal Transformer layers $N$ and the number of heads are empirically assigned to 4 and 8, respectively.
The unweighted average recall (UAR) and the weighted average recall (WAR) serve as the evaluation metrics.

\begin{table}[t]
	\caption{Comparison with other methods on FERV39K. The best results are in bold. }
	\centering
	\resizebox{0.4\textwidth}{!}{
	\begin{tabular}{c|c|cc}
		\toprule
		\multirow{2}{*}{Method} & \multirow{2}{*}{FLOPs}     & \multicolumn{2}{c}{Metrics (\%)}  \\
		\cmidrule{3-4}
		&  & UAR             & WAR            \\
	\midrule
	3D ResNet18              & -                                                                           & 26.67           & 37.57                                                                                \\
	ResNet18+LSTM          & -                                                                           & 30.92           & 42.59                                                                               \\
	VGG13+LSTM          & -                                                                           & 32.79           & 44.54                                                                               \\
	Former-DFER              & 9.11                                                                           & 37.20           & 46.85                                                                                 \\
	\midrule
	Baseline       & 25.16                                                                           & 37.41           & 47.63                                                                      \\           
	Baseline w/ Reg       & 25.16                                                                           & 37.76           & 48.11                                                                      \\           
	LOGO-Former w/o Reg       & 10.27                                                                           & 37.60           & 47.85                                                                      \\           
	\textbf{LOGO-Former}       & 10.27                                                                           & \textbf{38.22}           & \textbf{48.13}                                                                      \\           
\bottomrule
\end{tabular}
	}
\label{ferv39k}
\end{table}

\subsection{Comparison with State-of-the Art Methods}
We compare our method with other methods on DFEW and FERV39K, with respect to the UAR, the WAR.
The comparison result of DFEW and FERV39K are shown in Tab. \ref{dfew} and Tab. \ref{ferv39k}.
Our method obtains the best results using both metrics.
Specifically, Former-DFER is the previous state-of-the-art method with the UAR of 53.69\% and the WAR of 66.65\%.
Our method outperforms Former-DFER by 0.52\% and 1.28\% in terms of the UAR and the WAR, respectively.
Moreover, our method improves EC-STFL by 8.86\% and 10.47\% in UAR and WAR, which also aims to enhance the intra-class correlation and increase the inter-class distance.
As shown in Tab. \ref{dfew}, our method also obtains better results with respect to the category-level accuracy compared with other methods.

\begin{figure}[t]
	\centering
	\includegraphics[width=0.95\linewidth]{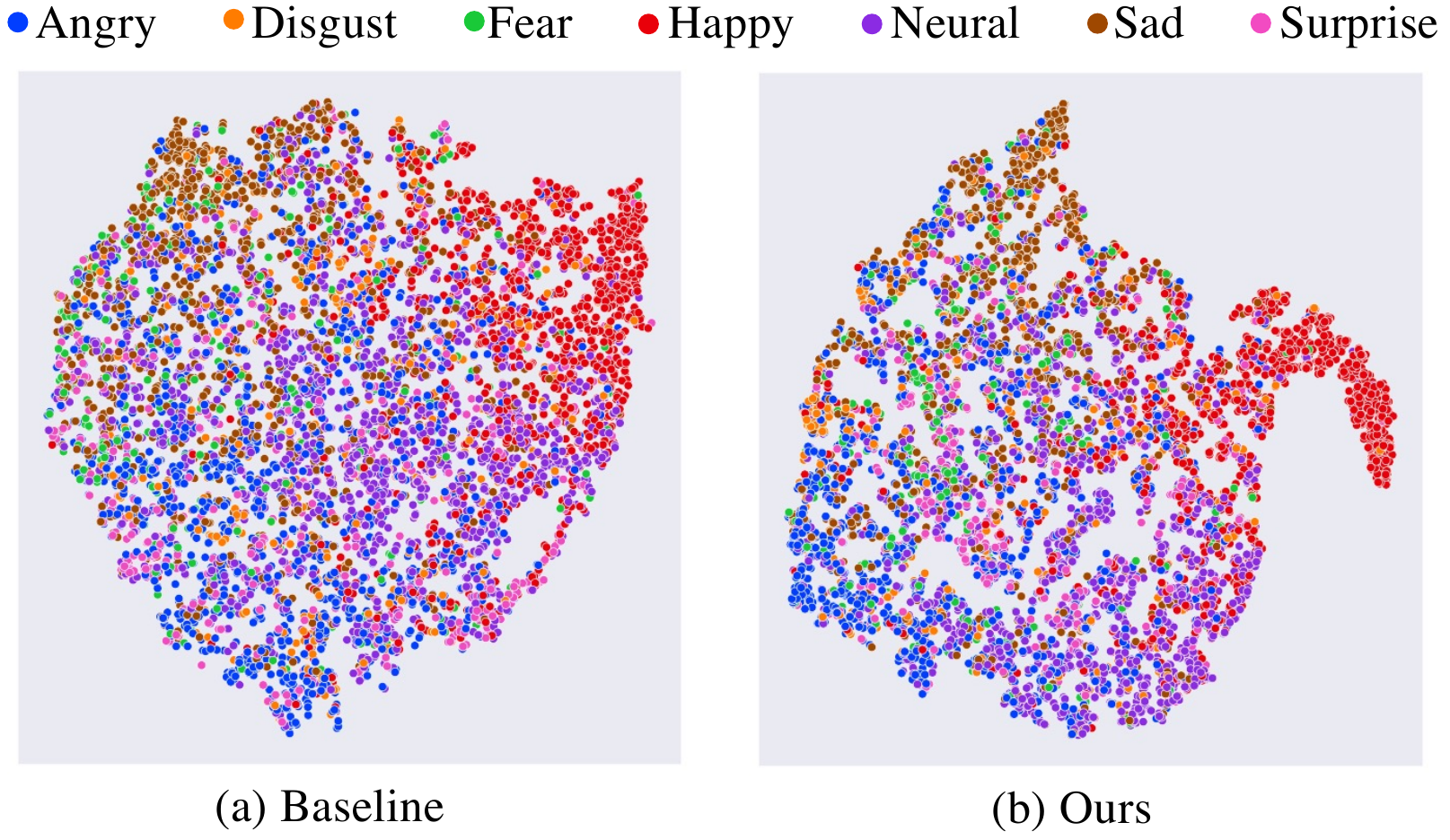}
	\caption{ Visualization of the feature distribution generated by t-SNE \cite{van2008visualizing} on FERV39K. (a) denotes our model with the standard cross entropy loss. (b) represents our model with the compact loss regularization
	term.}
	\label{tsne}
\end{figure}

We also conduct a further evaluation on FERV39K including FLOPs in Tab. \ref{ferv39k}.
We provide a strong baseline method by utilizing the vanilla spatial and temporal attention in \cite{ma2022spatio}.
The results in Tab. \ref{ferv39k} demonstrate that our model produces higher recognition performances while at the same time being significantly more efficient than our baseline method.
The visualization analysis in Fig. \ref{tsne} indicates our compact loss regularization term enables the learned features have a better aggregation effect and show more clear inter-class boundaries among different expressions.

\section{Conclusion}
\label{conclusion}
In this paper, we propose a simple but effective local-global Transformer (LOGO-Former) and the compact loss regularization term for in-the-wild dynamic facial expression recognition (DFER).
We jointly apply the local attention and the global attention within each block to learn spatio-temporal representations iteratively.
To further increase the model discriminant ability, we impose the constraint on the prediction distribution by the compact loss regularization term to enhance the intra-class correlation and increase the inter-class distance.
The experimental results and the visualization results demonstrate that our method learns discriminative spatio-temporal feature representations and enhances the classification margin.

\small
\bibliographystyle{IEEEbib}
\bibliography{strings,refs}

\end{document}